\documentclass[runningheads]{llncs}
\usepackage{graphicx}

\usepackage{tikz}
\usepackage{comment}
\usepackage{amsmath,amssymb} 
\usepackage{color}
\usepackage{url}
\usepackage{multirow}

\def\eg{{\em e.g.}}
\def\ie{{\em i.e.}}
\def\etal{{\em et al.}}

\begin{document}
\pagestyle{headings}
\mainmatter
\def\ECCVSubNumber{100}  

\title{VisDrone-CC2020: The Vision Meets Drone Crowd Counting Challenge Results} 

\titlerunning{VisDrone-CC2020}
%
\authorrunning{D. Du et al.}
%
\author{Dawei Du$^1$, Longyin Wen$^2$, Pengfei Zhu$^3$, Heng Fan$^4$, Qinghua Hu$^3$, Haibin Ling$^4$, Mubarak Shah$^5$, Junwen Pan$^3$, Ali Al-Ali$^{21}$, Amr Mohamed$^{12}$, Bakour Imene$^{14}$, Bin Dong$^{11}$, Binyu Zhang$^{13}$, Bouchali Hadia Nesma$^{14}$, Chenfeng Xu$^6$, Chenzhen Duan$^{20}$, Ciro Castiello$^{16}$, Corrado Mencar$^{16}$, Dingkang Liang$^6$, Florian Kr\"{u}ger$^9$, Gennaro Vessio$^{16}$, Giovanna Castellano$^{16}$, Jieru Wang$^8$, Junyu Gao$^7$, Khalid Abualsaud$^{12}$, Laihui Ding$^{15}$, Lei Zhao$^8$, Marco Cianciotta$^{16}$, Muhammad Saqib$^{18}$, Noor Almaadeed$^{12}$, Omar Elharrouss$^{12}$, Pei Lyu$^8$, Qi Wang$^7$, Shidong Liu$^{13}$, Shuang Qiu$^{17}$, Siyang Pan$^{13}$, Somaya Al-Maadeed$^{12}$, Sultan Daud Khan$^{19}$, Tamer Khattab$^{12}$, Tao Han$^7$, Thomas Golda$^{9,10}$, Wei Xu$^{13}$, Xiang Bai$^6$, Xiaoqing Xu$^{20}$, Xuelong Li$^7$, Yanyun Zhao$^{13}$, Ye Tian$^{20}$, Yingnan Lin$^8$, Yongchao Xu$^6$, Yuehan Yao$^{11}$, Zhenyu Xu$^{11}$, Zhijian Zhao$^{17}$, Zhipeng Luo$^{11}$, Zhiwei Wei$^{20}$, Zhiyuan Zhao$^7$}
\institute{
$^1$Kitware, Inc., Clifton Park, NY, USA.\\
$^2$JD Finance America Corporation, Mountain View, CA, USA. \\
$^3$Tianjin University, Tianjin, China.\\
$^4$Stony Brook University, New York, NY, USA.\\
$^5$University of Central Florida, Orlando, FL, USA. \\
$^6$Huazhong University of Science and Technology, Wuhan, China.\\
$^7$Northwestern Polytechnical University, Xi'an, China.\\
$^8$Unisoc, Inc., Shanghai, China.\\
$^9$Karlsruhe Institute of Technology, Karlsruhe, Germany.\\
$^{10}$Fraunhofer IOSB, Karlsruhe, Germany.\\
$^{11}$DeepBlue Technology(Shanghai) Co., Ltd, Shanghai, China.\\
$^{12}$Qatar University, Doha, Qatar.\\
$^{13}$Beijing University of Posts and Telecommunications, Beijing, China.\\
$^{14}$University Of Science And Technology Houari Boumediene, Bab Ezzouar, Algeria.\\
$^{15}$Ocean university of China, Qingdao, China.\\
$^{16}$University of Bari, Bari, Italy.\\
$^{17}$Supremind, Shanghai, China.\\
$^{18}$University of Technology Sydney, Ultimo, Australia.\\
$^{19}$National University of Technology, Islamabad, Pakistan.\\
$^{20}$Harbin Institute of Technology (Shenzhen), ShenZhen, China.\\
$^{21}$Supreme Committee for Delivery and Legacy, Doha, Qatar. \\
}
\maketitle

\begin{abstract}
Crowd counting on the drone platform is an interesting topic in computer vision, which brings new challenges such as small object inference, background clutter and wide viewpoint. However, there are few algorithms focusing on crowd counting on the drone-captured data due to the lack of comprehensive datasets. To this end, we collect a large-scale dataset and organize the Vision Meets Drone Crowd Counting Challenge (VisDrone-CC2020) in conjunction with the 16th European Conference on Computer Vision (ECCV 2020) to promote the developments in the related fields. The collected dataset is formed by $3,360$ images, including $2,460$ images for training, and $900$ images for testing. Specifically, we manually annotate persons with points in each video frame. There are $14$ algorithms from $15$ institutes submitted to the VisDrone-CC2020 Challenge. We provide a detailed analysis of the evaluation results and conclude the challenge. More information can be found at the website: \url{http://www.aiskyeye.com/}.
\end{abstract}

\section{Introduction}
Crowd counting aims to estimate the number of objects, \eg, pedestrians \cite{DBLP:conf/cvpr/ZhangZCGM16}, vehicles \cite{DBLP:conf/iccv/HsiehLH17}, commodity \cite{DBLP:journals/corr/abs-2003-08230}, animals \cite{DBLP:conf/eccv/ArtetaLZ16}, and cells \cite{DBLP:conf/nips/LempitskyZ10}, in images precisely. It has wide applications in video surveillance, crowd analysis, and traffic monitoring, to name a few.

With the developments of deep learning techniques in recent years, many researchers formulate the crowd counting problem as the regression of density maps using deep neural networks. For example, Zhang \etal \cite{DBLP:conf/cvpr/ZhangZCGM16} design a multi-column network architecture with three branches to deal with different scales of objects. After that, various algorithms \cite{DBLP:conf/cvpr/LiZC18,DBLP:conf/cvpr/LiuLZNPW19,DBLP:conf/iccv/ZhangSXZZ0019,DBLP:conf/cvpr/JiangXZZ0D019} achieve significant advances on several challenging datasets captured on surveillance scenes, such as UCF\_CC\_50 \cite{DBLP:conf/cvpr/IdreesSSS13}, WorldExpo \cite{DBLP:conf/cvpr/ZhangLWY15}, ShanghaiTech \cite{DBLP:conf/cvpr/ZhangZCGM16}, and UCF-QNRF \cite{DBLP:conf/eccv/IdreesTAZARS18}.

In contrast to images captured on surveillance scenarios, drone-captured video sequences involve difference challenges, including wide viewpoint variations, small objects and clutter background, which puts forward higher requirements of crowd counting algorithms. However, there are few datasets focus on such scenarios in the community. To advance the developments in crowd counting, we organize the Vision Meets Drone Crowd Counting (VisDrone-CC2020) challenge, which is one track of the ``Vision Meets Drone: A Challenge'' held on August 28, 2020, in conjunction with the 16th European Conference on Computer Vision (ECCV 2020). In particular, we provide a dataset, which is recorded by various drone-mounted cameras in $70$ different scenarios across $4$ different cities in China (\ie, Tianjin, Guangzhou, Daqing, and Hong Kong). The objects of interest are pedestrian. We invite researchers to submit the results of counting algorithms and share their latest research in the workshop. There are $14$ algorithms from $16$ institutes considered in the VisDrone-CC2020 Challenge. The detailed evaluation results can be found on the challenge website: \url{http://www.aiskyeye.com/}.

\section{Related Work}
In this section, we review the related crowd counting datasets and algorithms briefly. More details can be found in the survey \cite{DBLP:journals/corr/abs-2003-12783}.

\subsection{Crowd Counting Datasets}
Recently, numerous datasets have been proposed to deal with the challenges in crowd counting, such as scale variations, background clutter, and illumination variation in the wild. The most frequently used crowd counting datasets include
UCF\_CC\_50 \cite{DBLP:conf/cvpr/IdreesSSS13}, WorldExpo \cite{DBLP:conf/cvpr/ZhangLWY15}, ShanghaiTech \cite{DBLP:conf/cvpr/ZhangZCGM16}, and UCF-QNRF \cite{DBLP:conf/eccv/IdreesTAZARS18}. UCF\_CC\_50 \cite{DBLP:conf/cvpr/IdreesSSS13} contains only $50$ images from different scenes with various densities and different perspective distortions. WorldExpo \cite{DBLP:conf/cvpr/ZhangLWY15} is collected from videos of Shanghai 2010 WorldExpo, and include $3,920$ frames in total. ShanghaiTech \cite{DBLP:conf/cvpr/ZhangZCGM16} may be the most popular dataset (Part A and Part B) and composed of $1,198$ images with $330,165$ annotations. UCF-QNRF \cite{DBLP:conf/eccv/IdreesTAZARS18} is a large-scale high-resolution dataset that contains $1,535$ images and about $1.25$ million people heads.

However, the above-mentioned datasets are of relatively small size, which limits the power of deep learning. To avoid overfitting, recent proposed datasets collect more large-scale data in both the number of images and the number of persons, \eg, GCC \cite{DBLP:conf/cvpr/WangGL019} and Crowd Surveillance \cite{DBLP:conf/iccv/YanYZTWWD19}. The GCC dataset \cite{DBLP:conf/cvpr/WangGL019} is a diverse synthetic dataset collected from $400$ scenes in Grand Theft Auto V (GTA5). It includes $15,212$ images and $7,625,843$ persons, with resolution of $1080\times1920$. The Crowd Surveillance \cite{DBLP:conf/iccv/YanYZTWWD19} dataset contains $13,945$ high-resolution images and $386,513$ annotated people. Instead of surveillance scenes, in this work, we focus on crowd counting on the drone-captured scenes. Specifically, our proposed VisDrone-CC2020 dataset is formed by $3,360$ images, including $2,460$ images for training, and $900$ images for testing, which contains more than $400k$ annotated people heads. In Table \ref{fig:dataset}, we summarize the statistical comparison between our dataset and the previous datasets.

\subsection{Crowd Counting Methods}
The majority of early crowd counting methods \cite{DBLP:conf/iccv/WuN05,DBLP:conf/cvpr/LeibeSS05,DBLP:conf/cvpr/WangW11} rely on sliding-window detectors to scan still images or video frames to detect the pedestrians based on the hand-crafted features. However, these methods are easily affected by heavy occlusion, scale and viewpoint variations on crowded scenarios. Benefited from the great success of deep learning, many modern methods \cite{DBLP:conf/cvpr/ZhangZCGM16,DBLP:conf/cvpr/LiZC18} formulate crowd counting problem as regression of density maps by the networks. Zhang~\etal~\cite{DBLP:conf/cvpr/ZhangZCGM16} develop a multi-column architecture with three branches to deal with different scales of objects. To handle small objects, Li \etal~\cite{DBLP:conf/cvpr/LiZC18} employ dilated convolution layers to expand the receptive field while maintaining the resolution as backend network. In \cite{DBLP:conf/cvpr/LiuLZNPW19}, a multi-scale deformable network is proposed to generate high-quality density maps, which is more effective to capture the crowd features and more resistant to various noises. To capture interdependence of pixels in density maps, Zhang \etal~\cite{DBLP:conf/iccv/ZhangSXZZ0019} propose a Relational Attention Network with a self-attention mechanism. Jiang \etal~\cite{DBLP:conf/cvpr/JiangXZZ0D019} develop the trellis encoder-decoder network including a multi-scale encoder and a multi-path decoder to generate high-quality density estimation maps.

To achieve better performance, some recent methods exploit unlabeled or synthetic data for training. Based on the generated synthetic data \cite{DBLP:conf/cvpr/WangGL019}, the supervised learning and domain adaptation strategies are proposed to improve the counting accuracy significantly. In \cite{DBLP:journals/pami/LiuWB19}, the ranked image sets are generated from unlabeled data for counting applications suffering from a shortage of labeled data. Sam \etal~\cite{DBLP:conf/aaai/SamSMB19} propose the Grid Winner-Take-All autoencoder to learn features from unlabeled images such that weight update of neurons in convolutional output maps is restricted to the maximally activated neuron in a fixed spatial cell.

\section{The VisDrone-CC2020 Challenge}
The VisDrone-CC2020 Challenge aims to count pedestrian heads from video frames taken from drones. Participants are required to submit their algorithm and evaluate on the released VisDrone-CC2020 dataset. They are allowed to use external training data to improve the model. However, it is forbidden to submit different variants of the same algorithm. Meanwhile, the submission with detailed algorithm description obtains the authorship in the ECCV 2020 workshop proceeding.
\begin{table*}[t]
{
\centering
\caption{Comparison of existing crowd counting datasets. We summarize the maximal, minimal, average and total count in the datasets.}
\setlength{\tabcolsep}{3pt}
\begin{tabular}{c|ccccccc}
\hline
Dataset  &Type  &Frames &Max &Min &Ave &Total  &Year \\
\hline
UCSD \cite{DBLP:conf/cvpr/ChanLV08}               &surveillance     &$2,000$ &$46$          &$11$         &$24.9$       &$49,885$  &2008 \\
UCF\_CC\_50 \cite{DBLP:conf/cvpr/IdreesSSS13}    &surveillance   &$50$      &$4,543$      &$94$          &$1,279.5$  &$63,974$   &2013 \\
Mall \cite{DBLP:conf/iccv/LoyGX13}       &surveillance     &$2,000$ &$53$   &$13$       &$31.2$      &$62,316$ &2013 \\
WorldExpo  \cite{DBLP:conf/cvpr/ZhangLWY15}      &surveillance     &$3,980$ &$253$   &$1$            &$50.2$      &$199,923$ &2015 \\
Shanghaitech A \cite{DBLP:conf/cvpr/ZhangZCGM16} &surveillance &$482$    &$3,139$ &$33$    &$501.4$       &$241,677$   &2016 \\
Shanghaitech B \cite{DBLP:conf/cvpr/ZhangZCGM16} &surveillance &$716$ &$578$  &$9$   &$123.6$  &$88,488$  &2016 \\
AHU-Crowd \cite{DBLP:journals/jvcir/HuCNWL16}       &surveillance   &$107$ &$2,201$   &$58$    &$420.6$     &$45,000$ &2016 \\
CARPK \cite{DBLP:conf/iccv/HsiehLH17}      &aerial     &$1,448$ &$188$   &$1$            &$62.0$     &$89,777$ &2017 \\
Smart-City \cite{DBLP:conf/wacv/ZhangSC18}      &surveillance      &$50$ &$14$   &$1$            &$7.4$     &$369$ &2018 \\
UCF-QNRF \cite{DBLP:conf/eccv/IdreesTAZARS18}    &surveillance   &$1,535$   &$12,865$     &$49$      &$815.4$      &$1,251,642$  &2018 \\
FDST \cite{DBLP:conf/icmcs/FangZCGH19} &surveillance     &$15,000$ &$57$   &$9$            &$26.7$     & $394,081$ &2019 \\
GCC \cite{DBLP:conf/cvpr/WangGL019} &synthetic     &$15,212$ &$501$   &$0$            &$3995.0$     &$7,625,843$ &2019 \\
Crowd Surveillance \cite{DBLP:conf/iccv/YanYZTWWD19} &surveillance &$13,945$ &-   &-            &$35.0$     &$386,513$ &2019 \\
\hline
\hline
VisDrone-CC2020       &aerial  &$3,360$ &$421$ &$25$ &$144.7$ &$486,155$ &2020 \\
\hline
\end{tabular}}
\label{tab:dataset-comparison}
\end{table*}

\subsection{Dataset}
\begin{figure*}[t]
\centering
\includegraphics[width=1.0\linewidth]{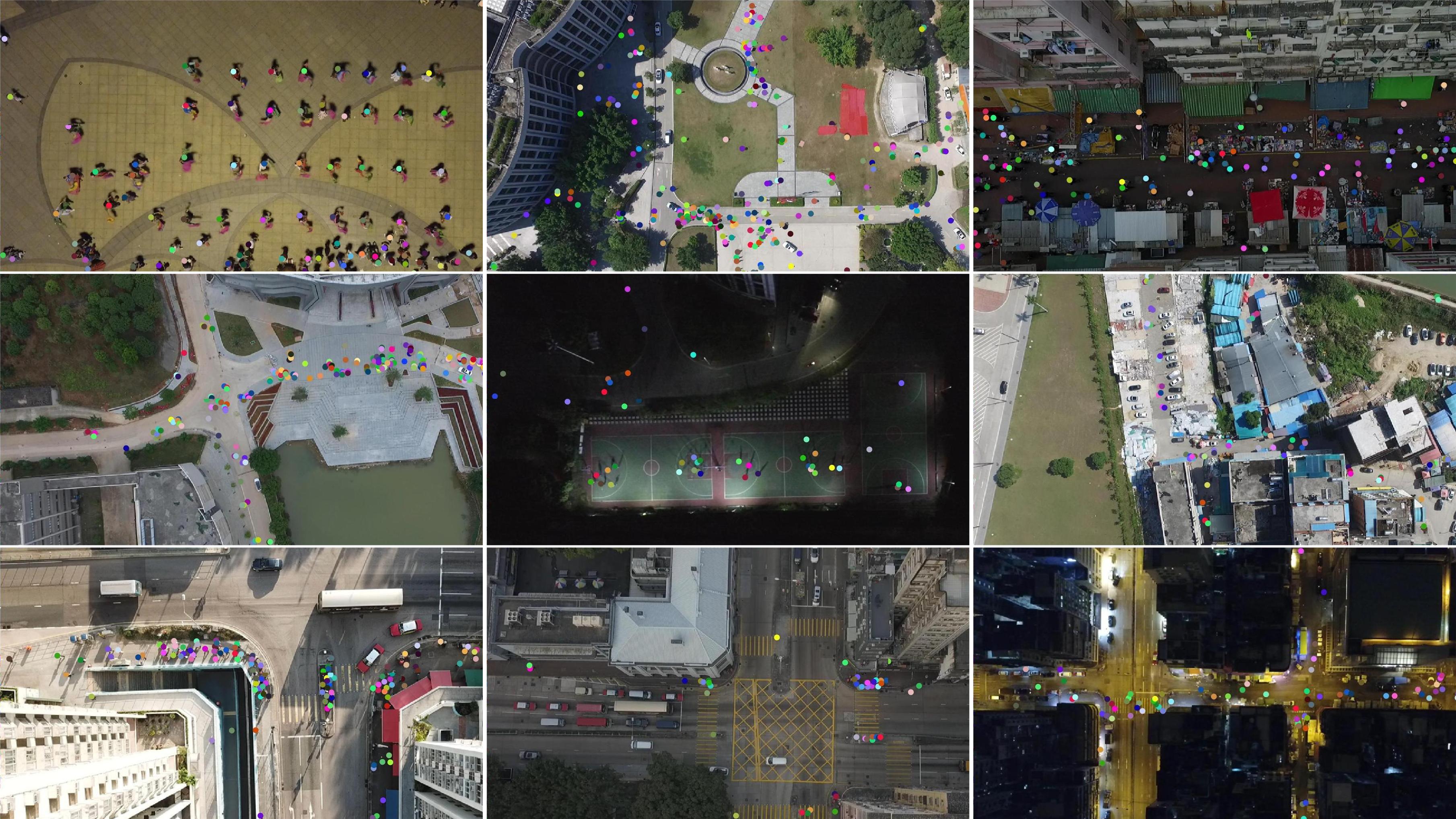}
\caption{Annotation exemplars in the VisDrone-CC2020 dataset. Different color indicates different person head.}
\label{fig:dataset}
\end{figure*}

The VisDrone-CC2020 dataset is formed by $3,360$ images with the resolution of $1920\times1080$. As shown in Fig. \ref{fig:dataset}, the data is captured by various drone-mounted cameras to keep diversity, for $70$ different scenarios across $4$ different cities in China (\ie, Tianjin, Guangzhou, Daqing, and Hong Kong). Moreover, we divide the dataset into the training and testing subsets, with $2,460$ and $900$ images, respectively. To avoid overfiting to particular scenes, we collect the images in the training and testing subsets at different but similar locations. To analyze the performance of algorithms thoroughly, we define $3$ visual attributes, described as follows.
\begin{itemize}
\item \textbf{Scale} indicates the size of objects. We define $2$ categories of scales including {\it Large} (the diameter of objects $>15$ pixels) and {\it Small} (the diameter of objects $\leq15$ pixels).
\item \textbf{Illumination} has significant influence on the appearance of objects. We define $3$ kinds of illumination conditions, \ie, {\it Cloudy}, {\it Sunny}, and {\it Night}.
\item \textbf{Density} is the number of objects in each frame. According to the average number of objects in each frame, we divide the dataset into $2$ density levels. {\it Crowded} density indicates that the number of objects in each frame is larger than $150$, and {\it Sparse} density indicates that the number of objects in each frame is less than $150$.
\end{itemize}

\subsubsection{Comparison to previous datasets.} Compared with the previous datasets focusing on crowd counting on surveillance scenes, our proposed dataset brings new challenges on drone-captured scenes as follows.
\begin{itemize}
\item Compared to objects in video sequences recorded on surveillance scenes, the scales of objects in our dataset are extremely small (even less than $30$ pixels) because of high shooting altitude by drones. It is difficult for the model to extract sufficient appearance information to describe the objects.
\item In our dataset, the crowds are scattered on video frames (see Fig. \ref{fig:dataset}). Each crowd contains a few to dozens of people.
\item Since the crowds are dynamically scattered on video frames, each crowd is surrounded by various backgrounds. The clutter background is another challenge in the proposed dataset.
\end{itemize}

\subsection{Evaluation Protocol}
Following the previous works \cite{DBLP:conf/cvpr/ZhangLWY15,DBLP:conf/cvpr/ZhangZCGM16}, each algorithm is evaluated through computing the number of people heads, mean absolute error (MAE) and mean squared error (MSE) between the predicted number of people heads and ground-truth in evaluation, which are defined as follows.
    \begin{equation}
    \label{eq:metric}
    \begin{aligned}
    &\text{MAE} = \frac{1}{\sum_{i=1}^{M}N_i}\sum_{i=1}^{M}\sum_{j=1}^{N_i}{|z_{i,j}-\hat{z}_{i,j}|},\\
    &\text{MSE} = \sqrt{\frac{1}{\sum_{i=1}^{M}N_i}\sum_{i=1}^{M}\sum_{j=1}^{N_i}{|z_{i,j}-\hat{z}_{i,j}|^2}},\\
    \end{aligned}
    \end{equation}
where $M$ is the number of video clips, $N_i$ is the number of frames in the $i$-th video. $z_{i,j}$ and $\hat{z}_{i,j}$ are the ground-truth and estimated number of people in the $j$-th frame of the $i$-th video clip, respectively. \text{MAE} and \text{MSE} describe the accuracy and robustness of the estimation, where \text{MAE} is the primary metric to rank the counting algorithms.

\section{Results and Analysis}
In this section, we evaluate the crowd counting methods submitted in the VisDrone-CC2020 Challenge and discuss the results thoroughly in terms of different attributes. Then, we point out several potential research direction in this field.

\subsection{Submitted Methods}
We have received $37$ entries in the VisDrone-CC2020 Challenge, $14$ of which submitted the results with correct format and complete algorithm description. In the following we brief overview the submitted algorithms included in the crowd counting task of VisDrone2020 Challenge and provide the corresponding descriptions in the Appendix \ref{submissions}.

The majority of the submitted algorithms are improved from state-of-the-art methods such as AutoScale \cite{DBLP:journals/corr/abs-1912-09632}, CSRNet \cite{DBLP:conf/cvpr/LiZC18} and SANet \cite{DBLP:conf/eccv/CaoWZS18}. FPNCC (\ref{app:fpncc}) is based on AutoScale \cite{DBLP:journals/corr/abs-1912-09632}. BVCC (\ref{app:bvcc}) is a double-stream network that extracts optical flow and frame difference information. $6$ algorithms are variants of CSRNet \cite{DBLP:conf/cvpr/LiZC18}, including PDCNN (\ref{app:pdcnn}), CSRNet+ (\ref{app:csrnet+}), SCNet (\ref{app:scnet}), CSR-SSOF (\ref{app:csr-ssof}) and Soft-CSRNET (\ref{app:soft-csrnet}). To extract multi-scale features of the target object and incorporate larger context, M-SFANet (\ref{app:m-sfanet}) improves SFANet \cite{DBLP:conf/icpr/ThanasutivesFNK20} by adding two modules called ASPP and CAN. MILLENNIUM (\ref{app:millennium}) uses multi-view data (\ie, real-world RGB image and the corresponding crowd heatmap) to construct two deep neural networks for crowd counting. DevaNetv2 (\ref{app:devanetv2}) employs attentional mechanism and feature pyramids to deal with different scales of people heads. SANet (\ref{app:sanet}) is a new encoder-decoder based Scale Aggregation Network \cite{DBLP:conf/eccv/CaoWZS18} to extract multi-scale features with scale aggregation modules and generate high-resolution density maps by using a set of transposed convolutions. Besides, two submissions are state-of-the-art methods trained on the VisDrone-CC2020 dataset, \ie, CFF (\ref{app:cff}) and CANet (\ref{app:canet}). CFF (\ref{app:cff}) proposes supervised focus from segmentation to focus on areas of interest and from global density to learn a matching global density. CANet (\ref{app:canet}) combines features obtained using multiple receptive field sizes and learns the importance of each such feature at each image location \cite{DBLP:conf/cvpr/LiuSF19}. ResNet-FPN101 (\ref{app:resnet}) is a baseline method by using ResNet-101 backbone to regress the density maps.
\begin{figure*}[t]
\centering
\includegraphics[width=1.0\linewidth]{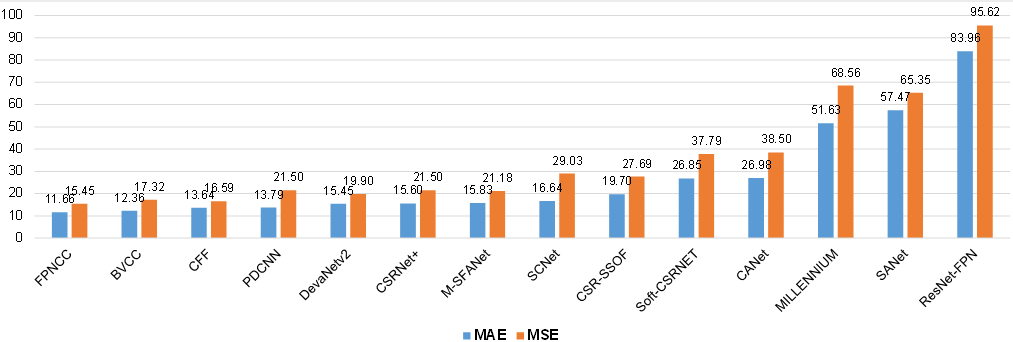}
\caption{Comparison of submissions in the VisDrone-CC2020 Challenge.}
\label{fig:res}
\end{figure*}

\subsection{Overall Results}
The overall results of all submissions are shown in Fig. \ref{fig:res}. FPNCC (\ref{app:fpncc}) obtains the best overall MAE score of $11.66$ and MSE score of $15.45$. This is attributed to the proposed Learning to Scale (L2S) module to rescale the dense regions into similar density levels, which mitigates imbalance of density values in the dataset. BVCC (\ref{app:bvcc}) ranks the second place by using a double-stream network, which introduces the external synthetic data generated by GTA5 \cite{DBLP:conf/cvpr/WangGL019}. After that, CFF (\ref{app:cff}) benefits from two kinds of point annotations (\ie, segmentation, global density) as supervision for density-based counting, achieving MAE score of $13.65$ and MSE score of $17.32$. PDCNN (\ref{app:pdcnn}) focuses on performing accurate counting under different illumination conditions, which obtains similar performance as CFF (\ref{app:cff}). After that, the following two methods focus on enhancing the training data. DevaNetv2 (\ref{app:devanetv2}) (rank 5) splits each image into $16$ sub-images, where each sub-image is processed by several data argumentation steps such as random rotation, random flip, random color process (include brightness, saturation, contrast), and normalization. Finally, $4$ sub-images are randomly chosen to merge into one new image. In constrast, CSRNet+ (\ref{app:csrnet+}) (rank 6) upsamples the training sequences to equalize the illumination distribution, which is pretrained on the DLR-Aerial Crowd Dataset \cite{DBLP:journals/corr/abs-1909-12743}.

\subsection{Attribute-based Results}
For thorough evaluation, we report the results in terms of different attributes in Table \ref{tab:density-map-estimation-results}. It can be seen that the best five performers achieve the best performance on various subsets. Specifically, improved from state-of-the-art L2S method \cite{DBLP:journals/corr/abs-1912-09632}, FPNCC (\ref{app:fpncc}) obtains the best MAE score in $4$ attribute subsets, \ie, \textit{Small}, \textit{Cloudy}, \textit{Sunny}, and \textit{Sparse}. With the help of the synthetic dataset \cite{DBLP:conf/cvpr/WangGL019} to simulate diverse environments, BVCC (\ref{app:bvcc}) achieves the best MSE score in $2$ attribute subsets including \textit{Cloudy} and \textit{Crowded}. PDCNN (\ref{app:pdcnn}) achieves the best MSE score in terms of \textit{Large} and \textit{Night} attributes. This is because PDCNN (\ref{app:pdcnn}) uses different networks to handle different scenarios in day and night illumination. DevaNetv2 (\ref{app:devanetv2}) obtains the best MAE score of $12.84$ in the \textit{Large} attribute, which show the effectiveness of the data argumentation strategy.

\begin{table*}[t]
\caption{Results on the VisDrone-CC2020 dataset.}
\centering
\setlength{\tabcolsep}{1pt}
\tiny{
\begin{tabular}{c||cc|cc||cc|cc|cc||cc|cc}
\hline
\multirow{2}{*}{Method} &\multicolumn{2}{c|}{Large} &\multicolumn{2}{c||}{Small} &\multicolumn{2}{c|}{Cloudy} &\multicolumn{2}{c|}{Sunny} &\multicolumn{2}{c||}{Night} &\multicolumn{2}{c|}{Crowded} &\multicolumn{2}{c}{Sparse}\\
\cline{2-15}
&$\text{MAE}$ &$\text{MSE}$ &$\text{MAE}$ &$\text{MSE}$ &$\text{MAE}$ &$\text{MSE}$ &$\text{MAE}$ &$\text{MSE}$ &$\text{MAE}$ &$\text{MSE}$ &$\text{MAE}$ &$\text{MSE}$ &$\text{MAE}$ &$\text{MSE}$\\
\hline
FPNCC(\ref{app:fpncc}) &13.74 &18.37 &\bf{10.27} &\bf{13.15} &\bf{10.83} &15.70 &\bf{11.04} &\bf{13.16} &15.39 &18.63 &15.40 &19.77 &\bf{9.49} &\bf{12.28}\\
BVCC(\ref{app:bvcc}) &13.48 &18.03 &11.61 &13.76 &11.78 &\bf{15.55} &12.32 &15.29 &14.19 &16.37 &\bf{13.66} &\bf{18.23} &11.61 &13.87\\
CFF(\ref{app:cff}) &13.73 &18.38 &13.58 &16.58 &13.28 &17.73 &13.35 &15.99 &15.32 &18.58 &18.69 &21.97 &10.72 &13.93 \\
PDCNN(\ref{app:pdcnn}) &13.01 &\bf{15.66} &14.32 &17.18 &14.84 &17.62 &13.43 &16.56 &\bf{11.36} &\bf{13.10} &15.79 &18.80 &12.63 &15.16\\
DevaNetv2(\ref{app:devanetv2}) &\bf{12.84} &15.84 &17.18 &22.19 &13.59 &17.97 &15.54 &18.66 &20.82 &26.65 &14.29 &17.35 &16.12 &21.23\\
CSRNet+(\ref{app:csrnet+}) &14.52 &20.38 &16.31 &22.21 &12.06 &17.55 &13.59 &16.91 &30.23 &35.75 &17.82 &22.70 &14.31 &20.78\\
M-SFANet(\ref{app:m-sfanet}) &20.59 &27.08 &12.65 &16.08 &15.56 &19.02 &11.22 &14.81 &25.88 &34.16 &17.24 &21.40 &15.01 &21.04\\
SCNet(\ref{app:scnet}) &19.64 &37.02 &14.64 &22.15 &13.77 &22.49 &11.53 &14.78 &35.47 &55.69 &18.40 &26.61 &15.62 &30.34\\
CSR-SSOF(\ref{app:csr-ssof}) &21.73 &30.23 &18.34 &25.86 &12.43 &18.53 &15.94 &19.66 &49.02 &52.90 &17.47 &24.23 &20.98 &29.51\\
Soft-CSRNET(\ref{app:soft-csrnet}) &29.31 &44.10 &25.21 &32.92 &15.37 &20.12 &19.73 &23.34 &75.53 &79.15 &21.86 &26.02 &29.74 &43.16\\
CANet(\ref{app:canet}) &30.92 &38.33 &24.36 &38.61 &11.97 &16.95 &23.40 &27.22 &79.18 &80.92 &24.06 &29.25 &28.67 &42.95 \\
MILLENNIUM(\ref{app:millennium}) &47.39 &53.70 &54.46 &76.89 &65.24 &85.26 &44.71 &51.83 &24.65 &32.00 &90.33 &103.25 &29.22 &35.36\\
SANet(\ref{app:sanet}) &56.81 &66.56 &57.91 &64.52 &44.77 &52.55 &69.91 &77.17 &70.67 &73.68 &70.14 &78.74 &50.13 &56.16\\
ResNet-FPN(\ref{app:resnet}) &98.64 &109.53 &74.18 &85.09 &80.94 &94.27 &84.51 &96.65 &91.95 &97.54 &131.11 &133.18 &56.67 &64.56\\
\hline
\end{tabular}}
\label{tab:density-map-estimation-results}
\end{table*}

\subsection{Discussion}
As presented in Table \ref{tab:density-map-estimation-results}, it can be seen that the best method FPNCC (\ref{app:fpncc}) achieves $11.66$ MAE score. That is, there are $8\%$ errors in average to count the people heads. It is still not satisfying in real applications. We summarize some topics worth to explore in crowd counting on drone-captured scenes as follows.
\begin{itemize}
\item \textbf{Groundtruth Density Map.} The majority of existing methods convert point based groundtruth to density map using a Gaussian kernel model training. Although the geometry-adaptive kernel-based density map generation \cite{DBLP:conf/cvpr/ZhangZCGM16} is widely used on surveillance scenes, it may fail on the drone-captured scenarios where the people crowd is relative sparse.

\item \textbf{Unsupervised Learning.} Since point-level annotation is expensive to collect, some methods leverage unlabeled data or synthetic external data to improve the crowd counting performance. To narrow the gap between different datasets, we believe the domain adaptation technique will attract much interest in the field.

\item \textbf{Head Localization.} Besides crowd counting, head localization is also an important task in safety control application. However, the submitted algorithms focus on estimating the number of people heads in a frame rather than accurate location. Previous works \cite{DBLP:conf/cvpr/MaYC15,DBLP:conf/eccv/IdreesTAZARS18,DBLP:journals/corr/abs-1912-01811} usually output the localization map and post-process the map by finding the local maximums based on a threshold. The two sub-tasks should be complementary and support each other.
\end{itemize}

\section{Conclusions}
In this paper, we summarize the results of all submitted crowd counting algorithms in the VisDrone-CC2020 Challenge. To evaluate the performance of algorithms, we provide a dataset formed by $3,360$ images, \ie, $2,460$ images for training, and $900$ images for testing. We provide annotated coordinates for people. Specifically, $14$ crowd counting algorithms from $15$ instittues are submitted to the VisDrone-CC2020 Challenge. The top three performers are FPNCC (\ref{app:fpncc}), BVCC (\ref{app:bvcc}), and CFF (\ref{app:cff}), achieving $11.66$, $12.36$ and $13.65$ MAE scores, respectively. However, there still remains much room for improvement such as the localization accuracy of head. For future work, we plan to extend the dataset with more attributes and annotations to advance the state-of-the-art. We hope our work can largely boost the development of crowd counting on the drone-captured scenes.

\section*{Acknowledgements}
This work was supported in part by the National Natural Science Foundation of China under Grant 61876127 and Grant 61732011, in part by Natural Science Foundation of Tianjin under Grant 17JCZDJC30800.

\appendix
\section{Submitted Crowd Counting Algorithms}\label{submissions}
In this appendix, we provide a short summary of all crowd counting algorithms that were considered in the VisDrone-CC2020 Challenge.

\subsection{Feature Pyramid Network for Crowd Counting (FPNCC)}
\label{app:fpncc}
\begin{figure*}[t]
\centering
\includegraphics[width=1.0\linewidth]{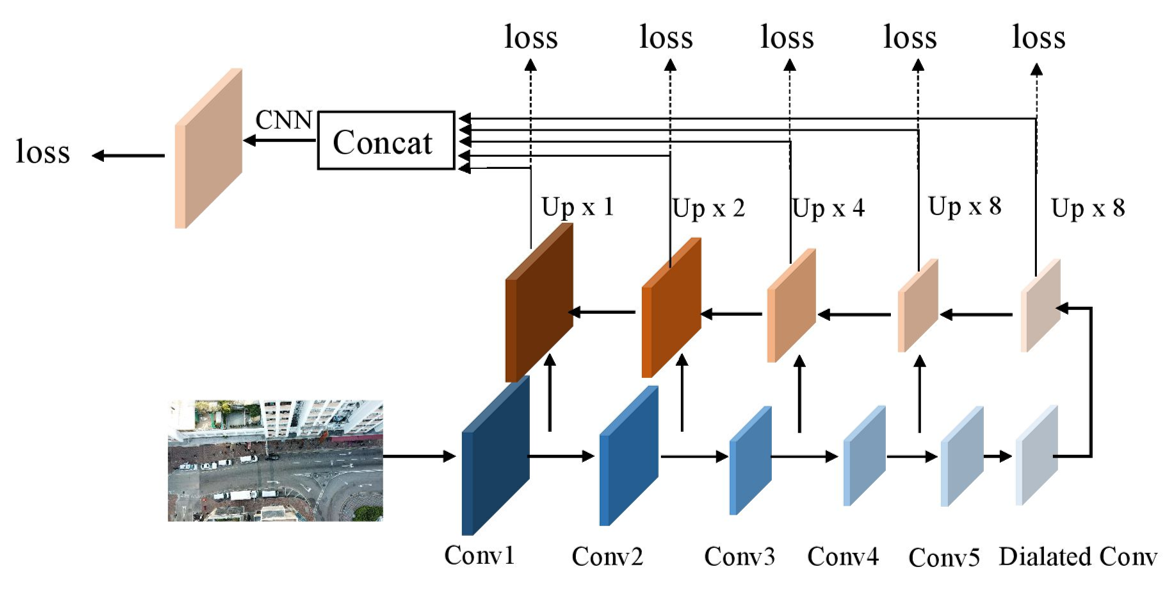}
\caption{The framework of FPNCC.}
\label{fig:fpncc}
\end{figure*}

\emph{Dingkang Liang, Chenfeng Xu, Yongchao Xu, Xiang Bai} \\
\emph{dkliang@hust.edu.cn} \\
\\
FPNCC is based on AutoScale \cite{DBLP:journals/corr/abs-1912-09632} (an extension of L2S~\cite{DBLP:conf/iccv/XuQFBXB19}) with the VGG16-based FPN backbone, which automatically scales dense regions into similar and appropriate density levels (see Fig. \ref{fig:fpncc}). Meanwhile, we separate the overlapped blobs and decompose the original accumulated density values in density map. To preserve sufficient spatial information for accurate counting, we discard the last pooling layer and all following fully connected layers, as well as the pooling layer between stage4 and stage5. Note that we exploit the pre-trained model based on ImageNet Database. Besides using the traditional MSE Loss, we also use SSIM loss to improve the high-quality density map to aid the final counting performance.

\subsection{Bi-Path Video Crowd Counting (BVCC)}
\label{app:bvcc}
\emph{Zhiyuan Zhao, Tao Han, Junyu Gao, Qi Wang, Xuelong Li} \\
\emph{tuzixini@163.com, hantao10200@mail.nwpu.edu.cn, \{gjy3035, crabwq\}@gmail.com, li@nwpu.edu.cn} \\
\\
To deal with the challenges such as varying scale, unstable exposure, and scene migration, BVCC is proposed to automatically understand the crowd from the visual data collected from drones. First, to alleviate the background noise generated in cross-scene testing, a double-stream crowd counting model is proposed, which extracts optical flow and frame difference information as an additional branch. Besides, to improve the generalization ability of the model at different scales and time, we randomly combine a variety of data transformation methods to simulate some unseen environments. To tackle the crowd density estimation problem under extreme dark environments, we introduce synthetic data generated by GTAV \cite{DBLP:conf/cvpr/WangGL019}.

\subsection{Counting with Focus for Free (CFF)}
\label{app:cff}
\emph{Wei Xu} \\
\emph{wxu.bupt@gmail.com} \\
\\
CFF \cite{DBLP:conf/iccv/ShiMS19} proposes two kinds of free supervision including segmentation maps and global density. Besides, an improved kernel size estimator is proposed to facilitate density estimation and the focus from segmentation. During training, we augment the images by randomly cropping $128\times128$ patches. Code is available at \url{https://github.com/shizenglin/Counting-with-Focus-for-Free}.

\subsection{Parallel Dilated Convolution Neural Network (PDCNN)}
\label{app:pdcnn}
\emph{Pei Lyu, Lei Zhao, Jieru Wang, Yingnan Lin} \\
\emph{\{pei.lv, michael.zhao, jieru.wang, lynn.lin\}@unisoc.com} \\
\\
PDCNN is the illumination-aware counting model based on CSRNet \cite{DBLP:conf/cvpr/LiZC18}. Since there are three categories of illumination conditions in the dataset (cloudy, sunny and night), we first distinguish the day and night illumination. Then, we use different networks to handle different scenarios. We use geometry-adaptive kernels to tackle the different congested scenes. We use Gaussian Kernel to blur each head annotation. We crop patches from each image at different location for data augmentation.

\subsection{Crowd Counting with Attentional Mechanism and Feature Pyramids (DevaNetv2)}
\label{app:devanetv2}
\emph{Ye Tian, Chenzhen Duan, Xiaoqing Xu, Zhiwei Wei} \\
\emph{\{19s151092, 18s151541, 19s051052, 19s051024\}@stu.hit.edu.cn} \\
\\
In the field of aerial image crowd counting, people usually use grid density maps as the label with ``unprejudiced'' neural network block, but there are a series of problems. On one hand, the scale and shape of people will change too much, according to the different shooting angles and flight heights of the UAV. On the other hand, the luminance of the image will change, because of the various shooting time. In order to solve the above problems, we use Gaussian density maps and useful data enhancement methods to improve the robustness of our model. We use attentional mechanism and feature pyramids to make model adjust the predicting results based on the global information. We also design an auxiliary loss to reduce the difficulty of model optimization. The network is implemented by the $C^3$ framework\footnote{\url{https://github.com/gjy3035/C-3-Framework}}.

\subsection{CSRNet on Drone-based Scenarios (CSRNet+)}
\label{app:csrnet+}
\emph{Florian Kr\"{u}ger, Thomas Golda} \\
\emph{\{florian.krueger, thomas.golda\}@iosb.fraunhofer.de} \\
\\
CSRNet+ is modified from CSRNet \cite{DBLP:conf/cvpr/LiZC18} to perform crowd counting on drone-based scenarios. We pretrain our model using only the training set from DLR-Aerial Crowd Dataset (DLRACD) \cite{DBLP:journals/corr/abs-1909-12743}. Then we finetune our model with the VisDrone2020-CC training dataset. For training on VisDrone2020-CC training dataset, we annotate the ground sampling distances (GSD) for the training set by hand and then generated density maps from those as in \cite{DBLP:journals/corr/abs-1909-12743}. Furthermore, we upsample the training sequences to equalize the illumination distribution. For night sequences, we only double them since there are only $3$ in the training set.

\subsection{ Mutil-Scale Aware based SFANet (M-SFANet)}
\label{app:m-sfanet}
\emph{Zhipeng Luo, Bin Dong, Yuehan Yao, Zhenyu Xu} \\
\emph{\{luozp, dongb, yaoyh, xuzy\}@blueai.com} \\
\\
We use M-SFANet \cite{DBLP:conf/icpr/ThanasutivesFNK20}, which modifies neural network architectures based on SFANet \cite{DBLP:journals/corr/abs-1902-01115} for crowd counting. Specifically, the method integrates two modules to improve the performance, \ie, atrous spatial pyramid pooling (ASPP) \cite{DBLP:journals/pami/ChenPKMY18} for multi-scale featurs and context-aware network (CAN) \cite{DBLP:conf/cvpr/LiuSF19} for corresponding contextual information. During training stage, the MSE loss is used for density map and cross entropy loss is used for attention map.
The whole dataset is split into $8$ folds, which means that $8$ independent models are trained. 
During inference stage, we use separated min clip to deal with videos in night scenes.
\subsection{Scaled Cascade Network for crowd counting on Drone data (SCNet)}
\label{app:scnet}
\emph{Omar Elharrouss, Noor Almaadeed, Khalid Abualsaud, Amr Mohamed, Tamer Khattab, Ali Al-Ali, Somaya Al-Maadeed} \\
\emph{elharrouss.omar@gmail.com} \\
\\
We propose a crowd counting method based on deep convolutional neural networks (CNN) by extracting high-level features to generate density maps that represent an estimation of the crowd count with respect to the scale variations in the scene. Specifically, SCNet is a CNN-based model by adding a cascade network after the frontend block inspired by those used in SPN \cite{DBLP:conf/wacv/ChenBSG19}, CSRNet \cite{DBLP:conf/cvpr/LiZC18}, and AGRD \cite{DBLP:conf/icassp/PanMZW20}. Cascade block is inspired by the Bi-directional cascade network in \cite{DBLP:conf/cvpr/HeZYSH19} used for edge detection. To handle the scale variations a Scale Enhancement Module (SEM) is introduced. The architecture employs sequential dilated convolution blocks with different kernels.

\subsection{CSRNet on Segmented Scenes with Optical Flow (CSR-SSOF)}
\label{app:csr-ssof}
\emph{Siyang Pan, Shidong Liu, Binyu Zhang, Yanyun Zhao} \\
\emph{\{pansiyang, lsd215, jl-lagrange, zyy\}@bupt.edu.cn} \\
\\
CSR-SSOF is derived from CSRNet \cite{DBLP:conf/cvpr/LiZC18} with several additional modules offering improvements. First, optical flow is computed based on the Gunnar Farneback algorithm \cite{DBLP:conf/scia/Farneback03} to extract temporal information. RGB images concatenated with corresponding optical flow are fed into the modified CSRNet which takes 5-channel data as input. Furthermore, we replace the last layer with the density map estimator (DME) in SANet \cite{DBLP:conf/eccv/CaoWZS18} in order to generate high-resolution density maps. The patch-based test scheme in SANet is also applied in our method. Considering the interference of irrelevant regions in which crowds rarely appear, an improved semantic segmentation model HRNetV2 \cite{DBLP:journals/corr/abs-1904-04514} is implemented to block out building areas. We extract the contours of segmentation maps following \cite{DBLP:journals/cvgip/SuzukiA85} and fill the small holes. The corrosion expansion morphology method is used to smooth and narrow down the boundaries of building areas. Besides, V channel of the HSV model is chosen for the division of day and night sequences and we take a series of measures to cope with the night ones. Night images are enhanced by RetinexNet \cite{DBLP:conf/bmvc/WeiWY018} at the beginning to make the buried details visible for counting. In addition, we assume buildings in night sequences exist in pixels with below-average V values, so we directly binarize the original night images accordingly to perform segmentation. All final counting results are obtained by the integral of density maps within non-building areas.
With partial parameters pre-trained on ImageNet, the modified CSRNet is finetuned on ShanghaiTech Part B \cite{DBLP:conf/cvpr/ZhangZCGM16} dataset and VisDrone2020 train set successively. For segmentation, HRNetV2 is firstly trained on UDD5 \cite{DBLP:conf/prcv/ChenWLCW18} dataset making use of Cityscapes \cite{DBLP:conf/cvpr/CordtsORREBFRS16} pre-trained weight. Then we manually annotate some images on VisDrone2020 train set distinguishing building areas and use the annotation to finetune our model. For the night image enhancement, we use RetinexNet model pre-trained on LOL \cite{DBLP:conf/bmvc/WeiWY018} dataset and RAISE \cite{DBLP:conf/mmsys/Dang-NguyenPCB15} dataset.

\subsection{Soft Dilated Convolutional Neural Networks for Understanding the Highly Congested Scenes (Soft-CSRNET)}
\label{app:soft-csrnet}
\emph{Bakour Imene, Bouchali Hadia Nesma} \\
\emph{\{imene.bakour, hadianesma.bouchali\}@etu.usthb.dz} \\
\\
Soft-CSRNET is modified on the network for Congested Scene Recognition called CSRNet \cite{DBLP:conf/cvpr/LiZC18} to provide a data-driven and deep learning method that can understand highly congested scenes and perform accurate count estimation as well as present high-quality density maps. We try several possible configurations on the network to be able to optimize it and render it in real time, for that we focus on the accuracy of the counting, removing the minimum possible resolution of the density maps estimated The proposed network is composed of two major components: a convolutional neural network (CNN) as the front-end for 2D feature extraction and a dilated CNN for the back-end, which uses dilated kernels to deliver larger reception fields and to replace pooling operations. Our Soft-CSRNET contains fewer parameters and convolutional layers in the backend.

\subsection{Context-Aware Crowd Counting (CANet)}
\label{app:canet}
\emph{Laihui Ding} \\
\emph{xmyyzy123@qq.com} \\
\\
CANet \cite{DBLP:conf/cvpr/LiuSF19} is an end-to-end network that leverages multiple receptive field scales to learn and combine feature at each image location. Thus the contextual information is encoded for predicting accurate crowd density.

\subsection{MultI-view fuLly convoLutional nEural Network for crowd couNting In drone-captUred iMages (MILLENNIUM)}
\label{app:millennium}
\emph{Giovanna Castellano, Ciro Castiello, Marco Cianciotta, Corrado Mencar, Gennaro Vessio} \\
\emph{\{giovanna.castellano, ciro.castiello, marco.cianciotta, corrado.mencar, }\\
\emph{gennaro.vessio\}@uniba.it} \\
\\
We couple multi-view data and multi-functional deep learning for efficient crowd counting in aerial images. Specifically, we exploit the real-world RGB image and the corresponding crowd heatmap as multiple views of the same scene containing a crowd to create a powerful regression model for crowd counting. Two deep neural networks, one for each view, are jointly trained so that their weights are updated at the same time. After training, only the network that processes the real-world images is retained as final model for crowd counting. This provides an accurate light-weight model that is suitable to meet the limited computational resources of a UAV. The final model is able to provide the crowd count with an average processing speed of about $125$ frames per second.

\subsection{Scale Aggregation Network (SANet)}
\label{app:sanet}
\emph{Shuang Qiu, Zhijian Zhao} \\
\emph{\{qiushuang, zhaozhijian\}@supremind.com} \\
\\
SANet is the encoder-decoder based Scale Aggregation Network \cite{DBLP:conf/eccv/CaoWZS18}. Specifically, the encoder is used to exploit multi-scale features by scale aggregation while the decoder is used to generate high-resolution density maps by transposed convolutions. To consider local correlation in density maps, both Euclidean loss and local pattern consistency loss are used to train the network for better performance. We finetune the network on additional datasets such as Shanghaitech B \cite{DBLP:conf/cvpr/ZhangZCGM16}, and Crowd surveillance \cite{DBLP:conf/iccv/YanYZTWWD19}.

\subsection{Residual Network Feature Pyramid Network\_101 (ResNet-FPN101)}
\label{app:resnet}
\emph{Muhammad Saqib, Sultan Daud Khan} \\
\emph{muhammad.saqib@uts.edu.au, sultandaud@gmail.com} \\
\\
ResNet-FPN101 is the baseline crowd counting method. Specifically, we convert the dot-level annotation to bounding box-level annotation. We train and evaluate the network with ResNet-FPN-101 as a backbone architecture.

%
%
\bibliographystyle{splncs04}
\bibliography{egbib}
\end{document}